\documentclass[sigconf,screen,nonacm]{acmart}
\setcopyright{none}
\renewcommand\footnotetextcopyrightpermission[1]{}

\AtBeginDocument{%
  }

\usepackage[normalem]{ulem}
\usepackage{subcaption}

\newcommand{\CUP}{CUP}
\newcommand{\press}{Crete University Press} 
\newcommand{\anonURL}{https://cup.gr/}
\renewcommand{\bibliofont}{\scriptsize}
\usepackage{microtype}
\usepackage{comment}
\usepackage{balance}

\begin{document}
\settopmatter{authorsperrow=3}
\title{A Comparative Evaluation of Embeddings and LLMs in a Greek Book Publisher Setting -- The \CUP\ Dataset}


\author{Katerina Papantoniou}
\affiliation{%
  \institution{ICS-FORTH}
  \city{Heraklion}
  \country{Greece}
}
\email{papanton@ics.forth.gr}

\author{Panagiotis Papadakos}
\affiliation{%
  \institution{ICS-FORTH}
  \city{Heraklion}
  \country{Greece}
}
\email{papadako@ics.forth.gr}

\author{Theodore Patkos}
\affiliation{%
  \institution{ICS-FORTH}
  \city{Heraklion}
  \country{Greece}
}
\email{patkos@ics.forth.gr}

\author{Dimitris Garefalakis}
\affiliation{%
  \institution{Crete University Press}
  \city{Heraklion}
  \country{Greece}
}
\email{garefalakis@cup.gr}

\author{Nikos Vardakis}
\affiliation{%
  \institution{Crete University Press}
  \city{Heraklion}
  \country{Greece}
}
\email{vardakis@cup.gr}

\author{Dimitris Plexousakis}
\affiliation{%
  \institution{ICS-FORTH}
  \city{Heraklion}
  \country{Greece}
}
\email{dp@ics.forth.gr}

\additionalaffiliation{%
\institution{University of Crete, Computer Science Department, Heraklion, Greece}
}

\setlength{\textfloatsep}{4pt}
\setlength{\floatsep}{4pt}
\setlength{\intextsep}{4pt}

\renewcommand{\shortauthors}{Papantoniou et al.}


\begin{abstract}
We present \CUP, a Greek book retrieval benchmark consisting of 868 catalog records and 104 expert-annotated queries with graded relevance judgments. We evaluate sparse (BM25), dense (sentence-transformers), hybrid, and LLM-assisted retrieval methods in this book-search setting. Multilingual embeddings outperform Greek-specific models, while hybrid retrieval performs best overall. A query-level analysis shows that BM25 excels at named-entity queries, while dense and hybrid methods improve natural-language, noisy, cross-lingual, and concept queries. Field-aware prompting has model-specific effects, while LLM TOC summarization improves TOC-only retrieval and LLM post-filtering improves early-stage retrieval at a high cost. Overall, \CUP\ enables  real-world evaluation of Greek retrieval across lexical, semantic, noisy, and cross-lingual queries.
\end{abstract}



\keywords{Greek information retrieval, multilingual embeddings, hybrid retrieval, large language models, book search, retrieval benchmark}


\maketitle
\begin{center}
\small\itshape
Preprint of a paper accepted for publication in the Proceedings of the 14th
EETN Conference on Artificial Intelligence (SETN 2026).
\end{center}

\section{Introduction}

Retrieval tasks in morphologically rich, relatively low-resource languages such as Greek remains challenging because large-scale benchmarks and domain-adapted models are scarce. The problem is particularly pronounced in book search, where queries range from short keywords to complex natural-language requests.

In this work, we focus on the catalog of the \press\ (\CUP)\footnote{\url{\anonURL}}, a diverse collection of academic and general-interest books spanning the sciences, humanities, and arts. Retrieval is complicated by Greek inflectional morphology, accents, lexical and semantic variation, domain-specific terminology, and heterogeneous metadata, including titles, descriptions, and tables of contents (TOCs).  Although TOCs contain valuable topical information, their sparse and structured form can limit their direct usefulness for retrieval.
To address these challenges, we introduce a curated benchmark for Greek book retrieval to evaluate modern embedding models and retrieval strategies. We use an LLM-based method to transform TOCs into summaries, and evaluate embedding models, hybrid retrieval, and LLM-based summarization and post-retrieval tasks. 

Our results highlight several key findings. First, strong multilingual embedding models consistently outperform Greek-specific ones, suggesting that scale and cross-lingual training remain critical for Greek-specific tasks. This gap is further amplified by the relative scarcity of Greek-specific models. Second, hybrid retrieval approaches improve performance over standalone sparse (lexical) or dense (semantic) methods. Third, LLM-based TOC summarization improves retrieval, while LLM-based post-filtering improves early precision at a high inference cost. 

This paper makes the following contributions: (a) we introduce a curated benchmark for Greek book retrieval with expert annotations\footnote{Dataset and detailed results at \url{https://gitlab.isl.ics.forth.gr/cup/cup-benchmark}}, (b) we conduct an evaluation of multilingual and Greek-specific embedding models, (c) we analyze hybrid retrieval strategies, (d) we evaluate LLM-based TOC summarization, and (e) we evaluate LLMs for post-retrieval filtering and reranking.

\section{Related Work}

\noindent{\textbf{Embedding models \& Greek}}
Sentence-transformer models such as SBERT made dense bi-encoder retrieval practical~\cite{reimers2019sentence-bert}, while multilingual models including BGE-M3, Qwen3, Nemotron, and Nomic have recently improved cross-lingual retrieval through larger training corpora, longer contexts, and instruction-aware representations
(see Table \ref{tab:embed-models}). For Greek, however, dedicated resources remain limited, with models such as Greek SBERT-based ~\cite{Zaikis2023dacl} and Greek-adapted XLM-R encoders \cite{papadopoulos2022farfetched} providing useful baselines.

\begin{table*}[t!]
\centering
\footnotesize
\begin{tabular}{lllllllll}
\textbf{Name} & \textbf{Year} & \textbf{Lang.} & \textbf{Word} & \textbf{Params} & \textbf{Max Tokens} & \textbf{Dim} & \textbf{Style} & \textbf{Ref.} \\ \midrule

nomic-embed-text-v2-moe (nomic) & 2025 & multi & subword & 432M (MoE) & 8192 & 768 & prefixed & \cite{nussbaum2025trainingsparsemixtureexperts} \\

qwen3-0.6B & 2025 & multi & subword & 0.6B & 32768 & 1024 & prompt & \cite{qwen3embedding2025}\\

qwen3-4B & 2025 & multi & subword & 4B & 32768 & 2560 & prompt & \cite{qwen3embedding2025}\\

llama-nemotron-embed-1b-v2 (nemotron) & 2024 & multi & subword & 1.1B & 8192 & 2048 & prefixed & \cite{moreira2024nv}\\

bge-m3 & 2024 & multi & subword & 1.1B & 8192 & 1024 & no prefix & \cite{bge-m3} \\

snowflake-arctic-embed-l-v2.0 (snowflake) & 2024 & multi & subword & 137M & 512 & 1024 & prefixed & \cite{yu2024arcticembed20multilingualretrieval}\\

mE5-large & 2024 & multi & subword & 560M & 512 & 1024 & prefixed & \cite{wang2024multilingual}\\

mE5-base & 2024 & multi & subword & 278M & 512 & 768 & prefixed & \cite{wang2024multilingual}\\

st-greek-media-bert-base-uncased (st-Greek-BERT) & 2023 & el & subword & 110M & 512 & 768 & no prefix & \cite{Zaikis2023dacl} \\

stsb-xlm-r-greek-transfer (xlm-r) & 2020 & el & subword & 550M & 512 & 768 & no prefix & \cite{papadopoulos2022farfetched} \\

distiluse-base-multilingual-cased-v2 (distiluse)& 2019 & multi & subword & 135M & 128 & 512 & no prefix & \cite{reimers2019sentence-bert}\\

paraphrase-multilingual-MiniLM-L12-v2 (minilm) & 2019 & multi & subword & 118M & 128 & 384 & no prefix & \cite{reimers2019sentence-bert} \\

paraphrase-multilingual-mpnet-base-v2 (mpnet) & 2019 & multi & subword & 125M & 128 & 768 & no prefix & \cite{reimers2019sentence-bert}\\


\bottomrule
\end{tabular}
\caption{Overview of evaluated subword embedding models, ordered by year.}
\vspace{-15pt}
\label{tab:embed-models}
\end{table*}

\noindent{\textbf{Retrieval-Augmented LLM Recommendation.}}
Recent recommendations approaches integrate retrieval mechanisms with generation.  BookGPT~\cite{li2023bookgpt} demonstrates how LLMs can model user preferences and generate personalized book recommendations by leveraging textual semantics. RAG-Rec~\cite{aly2026rag} combines retrieval-augmented generation with recommender systems, grounding outputs in user data and item metadata, reducing hallucinations and adapting to  dynamic user preferences.
More generally, \cite{wang2025building} outlines architectures where LLMs unify retrieval, ranking, and explanation.


\noindent{\textbf{LLMs and Greek.}}
LLMs still struggle in low-resource languages. Studies like \cite{tassios2025llm} emphasize the importance of model selection and adaptation for Greek in a chatbot that assists migrants.
GreekMMLU \cite{zhang2026greekmmlu}, comprising 21,805 multiple-choice
questions across 45 subjects, a benchmark for Greek language understanding, reveals performance gaps compared to high-resource settings, while GreekBarBench \cite{chlapanis2025greekbarbench} shows that LLMs underperform in the legal domain.   

To the best of our knowledge, there is currently no widely adopted benchmark for information retrieval in Greek. Suites that support Greek, like OYXOY \cite{kogkalidis2024oyxoy} and MMTEB \cite{enevoldsen2025mmteb} do not focus on retrieval performance. Multilingual retrieval benchmarks like mMARCO\cite{bonifacio2021mmarco}, BEIR\footnote{\url{https://github.com/beir-cellar/beir/wiki/Multilingual-datasets}}, and MIRACL\cite{zhang2023miracl}, do not support Greek.




\section{Dataset \& Benchmark}
The current \CUP\ book corpus dataset contains 868 books spanning four broad thematic categories: \textit{sciences (STEM)}, \textit{history and social sciences}, \textit{humanities} and \textit{arts}. While the vast majority of the corpus consists of Greek-language content, the catalog also contains a small portion of English-only titles, bilingual (Greek-English, and Greek-French) and trilingual. 
Each book record comprises fields such as \textit{title}, \textit{author}, \textit{categories}, \textit{tags},  \textit{ISBN}, that have a predefined schema and low format variability. It also contains fields such as \textit{content}, the publisher's description of the book and \textit{toc}, the raw table of contents, which consist of free-form natural language with variable length, inconsistent formatting, and heterogeneous semantic content. 
The descriptive content field averages approximately 225 words (max: 1,154), while the toc field exhibits similar average length (226 words) but considerably greater variance, ranging from empty to nearly 2,000 words. 342 books (39\% of the corpus) contain detailed tables of contents exceeding 200 words. The dataset also includes five LLM-generated summaries of TOCs (see Sec.~\ref{sec:eval}), used to enrich document representations for retrieval. 
Preprocessing included normalization of titles and author names, aggregation of categorical metadata, and removal of boilerplate from descriptions.

We constructed a custom ground-truth dataset tailored to the \CUP\ catalog. The dataset comprises 104 queries, each paired with manually curated relevance judgments over the corpus. Queries reflect diverse information needs and retrieval complexity. 
The query set spans short keyword queries (e.g., \textit{``Μανέτας''}) and natural language requests (e.g., \textit{``Με ενδιαφέρει να διαβάσω έργα του Γραμ\-μα\-τι\-κά\-κη\dots''}), enabling the evaluation  across varying linguistic complexity. It also includes multilingual queries expressed in English (e.g., \textit{``AI''}), reflecting cross-lingual search behavior. To further increase realism, the dataset includes orthographic variation, such as omitted diacritics and misspellings (e.g., \textit{``φεμισμός''} instead of \textit{``φεμινισμός''}), and queries with  constraints (e.g., \textit{``βιβλία φυσικής όχι του Τραχανά''}). The query set spans all major catalog categories.

Relevance judgments were performed manually by 4 domain experts familiar with the \CUP\ collection. For each query, the entire corpus was examined and books were assigned graded relevance labels based on how well they satisfy the underlying information need. A book was considered relevant if a reasonable user would find it useful for the query, with grading capturing varying degrees of usefulness. The query set includes single-word (30), multi-word (10), noisy (13), category-based (11), reasoning (11), natural language (10), other-language (9), and named-entity queries (10). The average number of relevant books per query is 4.81 (range 1–19).

\begin{table*}[htbp]
\setlength{\tabcolsep}{4pt}
\centering
\footnotesize
\begin{tabular}{l c c c c c c c c c c}
\toprule
\textbf{Model} & \textbf{Hits@1} & \textbf{Hits@3} & \textbf{Hits@9} & \textbf{Hits@20} & \textbf{nDCG} & \textbf{nDCG@3} & \textbf{nDCG@9} & \textbf{ndnDCGcg@20} & \textbf{MRR} & \textbf{R-Precision} \\
\midrule
\multicolumn{11}{c}{\textbf{Sparse (Lexical)}} \\
\midrule

okapi-BM25 & 0.567 & 0.721 & 0.788 & 0.808 & 0.622 & 0.537 & 0.544 & 0.567 & 0.652 & 0.452 \\
\midrule

\multicolumn{11}{c}{\textbf{Greek-Specific Dense (Semantic) Models. Values reported for (model/model*). Sorted by ndcg@9.}} \\
\midrule
xlm-r & \textbf{0.423}/0.385 & \textbf{0.519}/0.490	 & \textbf{0.577}/0.558 & 0.673/\textbf{0.683} & \textbf{0.518}/0.504 & \textbf{0.370}/0.350 & \textbf{0.372}/0.354 &\textbf{ 0.403}/0.388 & \textbf{0.487}/0.460 & \textbf{0.289}/0.258 \\

st-Greek-BERT & 0.192/0.202 & 0.346/0.375 & 0.500/0.471 & 0.606/0.558 & 0.379/0.375 & 0.178/0.181 & 0.203/0.201 & 0.226/0.222 & 0.305/0.304 & 0.155/0.151 \\
\midrule
\multicolumn{11}{c}{\textbf{Multilingual Dense (Semantic) Models. Values reported for (model/model*). Sorted by ndcg@9.}} \\
\midrule

nemotron & 0.606/\textbf{0.663} & 0.721/0.798 & 0.827/0.894 & 0.865/0.923 & 0.650/\textbf{0.690} & 0.517/0.555 & 0.534/\textbf{0.582} & 0.571/\textbf{0.619} & 0.684/\textbf{0.748} & 0.411/0.465 \\

nomic & 0.654/0.635 & 0.798/\textbf{0.817} & \textbf{0.904}/0.865 & 0.923/0.923 & 0.682/0.686 & \textbf{0.566}/0.561 & 0.567/0.577 & 0.605/0.617 & 0.745/0.735 & \textbf{0.473}/0.469 \\

bge-m3 & 0.596/0.606 & 0.712/0.788 & 0.875/0.894 & \textbf{0.933}/0.904 & 0.645/0.668 & 0.501/0.547 & 0.520/0.559 & 0.557/0.584 & 0.683/0.713 & 0.413/0.445 \\

qwen-4B (prompt-based) & 0.596 &0.788	 &0.856	 & 0.904 & 0.666 & 0.546 & 0.552 & 0.585 & 0.698 & 0.443 \\

e5-large & 0.596/0.529 & 0.721/0.750 & 0.846/0.885 & \textbf{0.933}/0.904 & 0.633/0.639 & 0.491/0.500 & 0.505/0.521 & 0.540/0.550 & 0.680/0.658 & 0.383/0.407 \\

snowflake & 0.577/0.577 & 0.740/0.740 & 0.894/0.894	 & 0.913/0.913 & 0.646/0.639 & 0.506/0.484 & 0.520/0.525 & 0.561/0.548 & 0.688/0.678 & 0.429/0.412 \\

e5-base & 0.452/0.529 & 0.673/0.673 & 0.798/0.779 & 0.827/0.827 & 0.561/0.585 & 0.395/0.436 & 0.424/0.443 & 0.455/0.482 & 0.575/0.617 & 0.313/0.365 \\

qwen0.6B (prompt-based) & 0.490 & 0.615 & 0.779 & 0.827 & 0.557 & 0.400 & 0.407 & 0.454 & 0.582 & 0.307 \\

mpnet & 0.433/0.404 & 0.529/0.519 & 0.587/0.615 & 0.683/0.673 & 0.509/0.503 & 0.374/0.357 & 0.363/0.361 & 0.390/0.385 & 0.494/0.482 & 0.271/0.269 \\

minilm & 0.385/0.375 & 0.510/0.481	 & 0.606/0.519 & 0.692/0.625 & 0.480/0.469 & 0.336/0.328 & 0.325/0.305 & 0.354/0.345 & 0.463/0.435 & 0.245/0.227 \\

distiluse & 0.279/0.298	 & 0.375/0.462	 & 0.500/0.538	 & 0.596/0.692& 0.413/0.450 & 0.232/0.280 & 0.245/0.290 & 0.270/0.323 & 0.357/0.400 & 0.168/0.205 \\

\midrule

\multicolumn{11}{c}{\textbf{Greedy Search Weighted  \& RRF Hybrid over sparse (s) and  top-2 dense (d1 = nemotron*, d2 = nomic*) models. Sorted by ndcg@9.}} \\
\midrule
\underline{s$_{0.4}$ + (d1$_{0.6}$ + d2$_{0.4}$)$_{0.6}$} & \textbf{0.740} & \textbf{0.865} & \underline{\textbf{0.942}} & 0.942  & \underline{\textbf{0.759}} & \textbf{0.658} & \underline{\textbf{0.673}} & \underline{\textbf{0.704}} & \textbf{0.810} & \underline{\textbf{0.568}} \\

s$_{0.45}$ + d1$_{0.55}$ &0.721 & 0.827 & 0.933 & 0.952 & 0.749 & 0.645 & 0.666 & 0.700 & 0.792 & 0.554  \\
s$_{0.5}$ + d2$_{0.5}$& 0.721 & 0.856	 &0.933&0.952& 0.747 & \textbf{0.658} & 0.658& 0.687 & 0.796&   0.543 \\

RRF(s + d1) & 0.654 & 0.817 & 0.913 & \underline{\textbf{0.962}} & 0.722 & 0.604 & 0.626 & 0.663 & 0.750 & 0.511  \\

RRF(s + d1 + d2) & 0.673 & 0.788 & 0.875 & 0.913 & 0.713 & 0.600 & 0.616 & 0.647 & 0.747 & 0.500\\

RRF(s + d2) & 0.683& 0.837		& 0.875	 & 0.913 & 0.723 & 0.623 & 0.622 & 0.656 & 0.763 & 0.531 \\

RRF(d1 + d2)  & 0.587	 & 0.683	 &0.769&  0.808 & 0.597 & 0.464 & 0.467 & 0.494 & 0.652 & 0.376\\
\midrule



\multicolumn{11}{c}{\textbf{Results over the LLM-based summarized TOC  field using s$_{0.4}$ + (d1$_{0.6}$ + d2$_{0.4}$)$_{0.6}$ hybrid model. Sorted by ndcg@9.}} \\
\midrule

EuroLLM-9B-Instruct-2512 & 0.558 & 0.712 & \textbf{0.827} & 0.865 & 0.621 & 0.510 & \textbf{0.508} & 0.530 & 0.652 & 0.383 \\

Meltemi-7B-Instruct-v1.5   & \textbf{0.587} & \textbf{0.750}  & 0.798	 & \textbf{0.913} & \textbf{0.628} & \textbf{0.513} & 0.502 & \textbf{0.539} & \textbf{0.679} & \textbf{0.405} \\

Llama-Krikri-8B-Instruct & 0.548 & 0.663 & 0.788 & 0.827& 0.608 & 0.472 & 0.495 & 0.528 & 0.623& 0.389 \\
Qwen3-8B & 0.519 & 0.654	 &0.779 & 0.817 & 0.586 & 0.470 & 0.475 & 0.486 & 0.606 & 0.368 \\

Deepseek-r1-qwen3-8b   & 0.577 & 0.673& 0.788 & 0.856 & 0.591 & 0.460 & 0.463 & 0.491 & 0.648 & 0.350 \\

base TOC   & 0.404 & 0.490 & 0.635 & 0.712 & 0.475 & 0.327 & 0.335 & 0.361 & 0.475 & 0.259 \\
\midrule

\multicolumn{11}{c}{\textbf{LLM-based post-filtering and reranking using 
s$_{0.4}$ + (d1$_{0.6}$ + d2$_{0.4}$)$_{0.6}$ hybrid model. Sorted by ndcg@9.}} \\
\midrule
Qwen3-8B  & \underline{\textbf{0.750}} & \underline{\textbf{0.885}} & \textbf{0.913} & - & \textbf{0.711} & \underline{\textbf{0.687}} & \textbf{0.666} & \textbf{0.650} & \underline{\textbf{0.821}} & \textbf{0.558} \\
DeepSeek-R1-Qwen3-8B  & 0.712 & 0.875 & 0.904 & - & 0.636 & 0.636 & 0.620 & 0.606 & 0.791 & 0.499 \\

EuroLLM-9B-Instruct-2512   & 0.692 & 0.827 & 0.865 & - & 0.626 & 0.632 & 0.598 & 0.584 & 0.765 & 0.505 \\
Llama-Krikri-8B-Instruct & 0.740 & 0.856 & 0.865 & - & 0.636 & 0.653 & 0.595 & 0.581 & 0.793 & 0.505 \\

Meltemi-7B-Instruct-v1.5   & 0.077 & 0.077 & 0.077 & - & 0.060 & 0.052 & 0.043 & 0.042 & 0.077 & 0.034 \\ 

\bottomrule
\end{tabular}
\caption{Results for sparse, dense models, hybrid fusion methods, and LLM-based summarization and reranking.}
\label{tab:results}
\vspace{-20pt}
\end{table*}

\section{Evaluation}
\label{sec:eval}

We construct sparse and dense representations over five key fields: \textit{title}, \textit{author}, \textit{categories+tags}, \textit{content}, and \textit{toc}, all with equal weights.
TOCs are summarized using five LLMs (see below), and we use the \textit{krikri} ones for the best method retrieval evaluation.
For lexical retrieval, we employ a BM25-based retriever using the \texttt{rank\_bm25} library
integrated within the LangChain framework.
Standard normalization techniques are applied, including lowercasing, accent stripping, stemming, and stopword removal. 
We evaluate 13 embedding models using Sentence-transformers
(see Table \ref{tab:embed-models}),  following the recommended prompting or formatting conventions for each model. We also explore 9 variations (marked with a \textit{model*}) with field-aware prefixes (e.g., \textit{``Ο συγγραφέας του βιβλίου είναι \dots''} for the \textit{author} field) to better capture field semantics. The resulting embeddings are indexed using FAISS
for efficient similarity search.
We evaluate hybrid strategies that combine lexical and semantic signals, and extend them to a multi-dense setting with the top two dense models. The weights of sparse and dense indices are tuned via greedy search.  We also consider a \textit{Reciprocal Rank Fusion (RRF)} strategy. For each query, books are assigned fused scores based on the reciprocal of their ranks across multiple models to produce a top-$k$ list, where $\mathrm{RRF}(b) = \sum_{r \in R} \frac{1}{k + r(b)}$ for book $b$, the set of rankers $R$, the smoothing constant $k = 60$, and $r(b)$ the rank of $b$ produced by $r$. 
We evaluate the Greek-focused
\textit{Meltemi-7B}\footnote{\url{https://www.ilsp.gr/en/news/meltemi-en/}} and \textit{Llama-Krikri-8B}~\cite{roussis2025krikri}, and the multilingual \textit{EuroLLM-9B}~\cite{Martins2025EuroLLM},
\textit{Qwen3-8B}~\cite{yang2025qwen3} , and \textit{DeepSeek-R1-Qwen3-8B}~\cite{guo2025deepseek} LLMs.

We report standard IR metrics: \textit{nDCG@\{9,20\footnote{Due to GPU context limits (AMD Radeon AI PRO R9700 32GB and model restrictions), we do not report results at $k=20$ for LLM filtering and reranking.}\}}, \textit{MRR}, \textit{Hits@\{1,3,9\}}, and \textit{R-Precision}. nDCG measures ranking quality under graded relevance, MRR captures early precision, and Hits@k measures the fraction of queries with at least one relavent book in the top-$k$ results. Small $k$ values correspond to limited display settings, while large ones evaluate deeper ranking. Metrics are computed using \texttt{rank-eval}~\cite{bassani2022ranx}, a \texttt{trec\_eval}-compatible Python library.

Table~\ref{tab:results} reports retrieval performance, with best-in-category results shown in bold and the overall best underlined. The detailed results, including per category performance and an ablation study\footnote{Ablation shows that removing any field degrades performance, with the largest drop without the author field (0.606 vs.\ 0.675 nDCG@9), and the smallest for categories.} of the importance of each field is available in the online repository. Overall, multilingual dense models substantially outperform the sparse \textit{BM25} baseline. The strongest standalone dense model is \textit{nemotron*}, which reaches 0.582 nDCG@9, 0.619 nDCG@20, and 0.748 MRR, compared to 0.544, 0.567, and 0.652 for \textit{BM25}. 
Among dense models, multilingual embeddings clearly outperform Greek-specific encoders. The best Greek model (\textit{xlm-r}) reaches only 0.372 nDCG@9, while several multilingual models exceed 0.50.
Field-aware prefixes improve some models, notably \textit{nemotron} (nDCG@9: 0.534 $\rightarrow$ 0.582, MRR: 0.684 $\rightarrow$ 0.748), with smaller gains for \textit{bge-m3}, but is inconsistent across others, indicating model dependence. The best results are obtained by weighted hybrid retrieval, where the $s_{0.4} + (d1_{0.6} + d2_{0.4})_{0.6}$ performs best (0.673 nDCG@9, 0.810 MRR), improving over the best dense model (0.582, 0.748) and confirming complementarity between lexical and semantic cues. \textit{RRF} performs worse, indicating score-level weighting effectiveness over rank-based fusion. The best hybrid method gains are statistically significant over most metrics ($p<0.01$), except for MRR vs.\ \textit{nemotron*}.

The per-query-category results further clarify these gains. \textit{BM25} remains extremely strong for exact named-entity queries, particularly authors, where it reaches perfect nDCG@9. Dense and hybrid methods provide substantial improvements for query types requiring semantic understanding or tolerance to surface variation. For natural-language queries, nDCG@9 improves from 0.374 (\textit{BM25}) to 0.553 (best hybrid), and for noisy queries from 0.488 to 0.672, highlighting tolerance to misspellings and missing diacritics. Hybrid retrieval improves title and concept queries. It achieves perfect performance on named-title queries (vs.\ 0.765 \textit{BM25}) and outperforms both \textit{BM25} and dense models on concept queries. Dense-only retrieval can be preferable in cross-lingual and reasoning queries,
with \textit{nemotron*} leading nDCG@9.
Six noisy, natural language, multiword concept, and category queries yield no relevant books in the top-9 results across all configurations.

The TOC-only experiment shows that LLM-generated summaries ($\sim 500$ words) improve retrieval over raw TOC indexing (prompts available in the repository), with lower length variance but reduced lexical diversity. \textit{deepseek} is more verbose, while \textit{krikri} and \textit{euroLLM} are more concise. \textit{meltemi} has high scores since it fails to follow the prompt, leaking contextual metadata (author, title, description) into the summary. \textit{euroLLM} achieves the best nDCG@9 and hits@\{3,9\}, which alongside with \textit{krikri} produce the most linguistically correct Greek summaries. Overall, summary quality and prompt adherence significantly affect retrieval performance. 

For LLM-based post-filtering, models were given the top-9 results from the best hybrid retriever with full metadata and were asked to select and rank relevant items. \textit{qwen3} performs best overall, improving Hits@1, nDCG@3, and MRR (0.821), but returns shorter filtered lists, lowering Hits@9 relative to retrieval-only systems. \textit{deepseek}, \textit{euroLLM}, and \textit{krikri} are worse than the best hybrid, while \textit{meltemi} fails due to poor prompt adherence. Invalid or missing ID rates were 75\% for \textit{meltemi}, 3.85\% for \textit{euroLLM}/\textit{krikri}, 2.88\% for \textit{deepseek}, and 1.92\% for \textit{qwen3}. The  \textit{qwen3} gains come at a very high per-query inference cost in our setting (avg 92.5 secs).

Overall, no single retrieval strategy dominates across all query types. BM25 is the strongest for exact-match queries, dense models excel in reasoning and cross-lingual scenarios, and hybrid retrieval provides the best overall performance. LLM-based components offer complementary gains, with summarization enriching representations and post-filtering improving early precision, however at a higher computational cost for a real-time setting.

\section{Conclusion and Future Work}
We introduced \CUP, a Greek book retrieval benchmark with 868 records and 104 expert-annotated queries, and evaluated sparse, dense, hybrid, and LLM-based methods. Multilingual embeddings outperform Greek-specific models, while weighted hybrid retrieval performs best (0.673 nDCG@9), highlighting lexical and semantic complementarity. BM25 excels at exact named-entity queries, while dense and hybrid methods improve natural-language, noisy, reasoning, cross-lingual, and concept queries. LLM-based summarization enriches sparse TOC fields, while LLM post-filtering improves early-stage retrieval but is costly for real-time use. Future work will extend the benchmark and further evaluate LLM-based reranking and answer generation in interactive settings.

\balance

\bibliographystyle{ACM-Reference-Format}
\bibliography{references}
\end{document}